\documentclass[letterpaper]{article} 
\usepackage[draft]{aaai2026}  
\usepackage{times}  
\usepackage{helvet}  
\usepackage{courier}  
\usepackage[hyphens]{url}  
\usepackage{graphicx} 
\usepackage{natbib}  
\usepackage{caption} 
\usepackage{colortbl}
\usepackage{xcolor}
\usepackage{multirow}
\usepackage{algorithm}
\usepackage{algorithmic}
\usepackage{amsmath}

\usepackage{booktabs}
\usepackage[utf8]{inputenc}

\usepackage{wasysym} 
\DeclareUnicodeCharacter{266B}{\twonotes} 

\usepackage{newfloat}
\usepackage{listings}
\DeclareCaptionStyle{ruled}{labelfont=normalfont,labelsep=colon,strut=off} 
\floatstyle{ruled}
\newfloat{listing}{tb}{lst}{}
\floatname{listing}{Listing}
\title{Transforming Questions and Documents for Semantically Aligned Retrieval-Augmented Generation}
\author{
    Seokgi Lee
}
\affiliations{

    seokilee@snu.ac.kr
}

\begin{document}

\maketitle

\begin{abstract}
We introduce a novel retrieval‑augmented generation (RAG) framework tailored for multihop question answering. First, our system uses large language model (LLM) to decompose complex multihop questions into a sequence of single‑hop subquestions that guide document retrieval. This decomposition mitigates the ambiguity inherent in multihop queries by clearly targeting distinct knowledge facets. Second, instead of embedding raw or chunked documents directly, we generate answerable questions from each document chunk using Qwen3‑8B, embed these generated questions, and retrieve relevant chunks via question–question embedding similarity. During inference, the retrieved chunks are then fed along with the original question into the RAG pipeline. We evaluate on three multihop question datasets (MuSiQue, 2WikiMultiHopQa, HotpotQA) from LongBench. Our method improves RAG performacne compared to baseline systems. Our contributions highlight the benefits of using answerable‑question embeddings for RAG, and the effectiveness of LLM‑based query decomposition for multihop scenarios.
\end{abstract}

\section{Introduction}

Large language models (LLMs) are impressive in what they can do, but they’re not perfect—especially when it comes to answering questions that require external facts. Retrieval-Augmented Generation (RAG) tries to close this gap by bringing in relevant documents during inference. Since it was first introduced by ~\cite{lewis2020retrieval}, RAG has been used in all sorts of tasks like open-domain question answering, fact-checking, and summarization, where grounding responses in real-world information really matters.

But there’s a catch: most RAG systems still rely on embedding the entire user question into a single dense vector and using that to retrieve documents. This approach works reasonably well for straightforward questions. But when it comes to multi-hop questions—those that involve piecing together facts from multiple sources—it often falls short. The model may retrieve something vaguely related but miss key steps along the way.

In response to this challenge, various studies have explored structured strategies for improving retrieval. For example, some methods use graphs to represent how pieces of information are connected across documents—such as G-RAG~\cite{dong2024don}, CUE-RAG~\cite{su2025cue}, and HLG~\cite{ghassel2025hierarchical}. Others, such as HiRAG and DeepRAG~\cite{zhang2024hierarchical, ji2025deeprag}, use multi-layered retrieval pipelines that narrow down the search space in stages.

These systems bring real gains, but they also bring complexity. We were interested in whether a simpler approach could work just as well. One promising idea is to break down a complex question into smaller, more focused subquestions. This makes it easier to retrieve relevant information for each step and gives the model a clearer path to the answer.

That’s where our method comes in. Instead of trying to handle the full complexity of a multi-hop question all at once, we decompose it into single-hop subquestions, retrieve evidence for each, and then use a reranking model to choose the passages that best match the original question. It’s a way to stay focused on what matters, without over-engineering the pipeline.

Still, even with decomposition and reranking, there's an important issue we had to address: the format mismatch between user queries and document content. Questions are usually short, focused, and interrogative. Documents, on the other hand, are longer, narrative, and descriptive\cite{furnas1987vocabulary}. Even if the content is semantically aligned, the structural differences often confuse embedding-based retrieval systems~\cite{zhao2010term, zahhar2024leveraging}.

To address this mismatch, we take a different approach: instead of forcing a direct match between question and document as-is, we reshape both. On the query side, we decompose the complex multi-hop question into single-hop parts. On the document side, we generate answerable questions (AQs) from each chunk using a large language model.

This isn’t just a structural trick—it enriches the document representation itself. For an LLM to ask meaningful, answerable questions about a chunk of text, it has to process that chunk from multiple angles: what's stated explicitly, what’s implied, and what could reasonably be asked about. In doing so, it effectively distills and reorganizes the content in a way that makes it more semantically accessible for retrieval. The result is a document that becomes not just a source of information, but a map of the questions it can answer.

To address this, we propose a question-centric RAG framework that aligns both sides of the retrieval process through question formatting. Our main contributions are as follows:

\begin{itemize}
    \item We present a novel RAG pipeline for multihop question answering that leverages LLM-based question decomposition to transform complex questions into semantically focused single-hop subquestions, enabling more accurate and targeted retrieval without requiring task-specific supervision.
    
    \item We replace standard chunk embeddings with answerable question (AQ) representations, generated from each document chunk using Qwen3-8B~\cite{yang2025qwen3} without any fine-tuning. This approach aligns the semantic and syntactic structure of document and query embeddings, reducing structural mismatch and improving retrieval quality through question-to-question similarity.
    
    \item Our method requires no online multistage summarization, paraphrasing, or graph construction, yet achieves consistently strong performance across three multihop QA benchmarks—HotpotQA~\cite{yang2018HotpotQA}, 2WikiMultiHopQa~\cite{ho2020constructing}, and MuSiQue~\cite{trivedi2022MuSiQue}—in answer accuracy.
\end{itemize}

\section{Related Works}

\subsection{RAG for Multihop Question Answering}
Retrieval-Augmented Generation (RAG)~\cite{lewis2020retrieval} was originally designed for answering single-hop factual questions. It has since been extended to multihop scenarios through techniques such as Fusion-in-Decoder~\cite{izacard2020leveraging} and multihop retriever adaptation~\cite{asai2019learning}, which aggregate multiple passages for reasoning. Despite these enhancements, most approaches rely on static dense embeddings of document chunks, limiting their ability to capture multi-step dependencies and reasoning paths.

\subsection{Handling Multi-step Reasoning in Retrieval}
Graph-based methods such as G-RAG~\cite{dong2024don} and CUE-RAG~\cite{su2025cue} use semantic relationships between documents to support multihop retrieval. Hierarchical frameworks like HiRAG~\cite{ghassel2025hierarchical} and DeepRAG~\cite{ji2025deeprag} refine relevance through layered filtering. Recent advances include LongRAG~\cite{jiang2024longrag}, which handles long-document contexts, and MacRAG~\cite{lim2025macrag}, which leverages memory-augmented compositional reasoning. These methods typically introduce additional architectural components such as graph traversal or staged retrieval pipelines.

\subsection{Structural Mismatch Between Queries and Documents}
A key limitation in embedding-based retrieval lies in the structural discrepancy between queries and documents. Queries are typically short and interrogative, while documents are longer and descriptive. This mismatch leads to low embedding similarity despite semantic alignment~\cite{zhao2010term, zahhar2024leveraging}. To mitigate this, some works explore query rewriting~\cite{ma2023query}, document summarization~\cite{jiang2025retrieve}, or query paraphrasing~\cite{dong2017learning}.While these approaches adjust the surface form of queries or documents—or enhance them with greater informational diversity—they do not fundamentally restructure documents to align with the logic of question answering.

\subsection{Question-Centric Document Transformation}
Despite significant progress in retrieval-augmented generation, many existing systems continue to treat documents as static inputs, relying on surface-level matching. Our work takes a different approach by reformulating document chunks into answerable questions (AQs) using large language models. This process not only aligns the structure and semantics of documents with natural query forms, but also reconfigures the document content itself---capturing what is explicitly stated, as well as what is implied or inferable. By distilling documents into question-oriented representations, our method enriches their semantic accessibility and improves their utility for multi-hop reasoning. This dual transformation---query decomposition and AQ generation---constitutes, to our knowledge, the first unified framework that structurally aligns both sides of the RAG pipeline in a question-centric format.

\section{Method}

\subsection{Overview}
We introduce a novel RAG framework that employs question-centric transformations to align both the multihop question and the document structurally, thereby enhancing retrieval-augmented generation. This framework comprises four key stages:
\begin{itemize}
    \item \textbf{(1) Question Decomposition with LLMs.} Multihop questions are decomposed into single-hop subquestions using LLM. This decomposition is designed to semantically isolate each sub-question, facilitating more direct and effective retrieval.
    \item \textbf{(2) Document-side Answerable Question Generation (AQG).} For each document chunk, we apply pretrained Qwen3-8B to generate a set of answerable questions. These questions serve as semantic proxies for the document in the same embedding space as user queries.
    \item \textbf{(3) Two-Stage Retrieval and Reranking.} Each decomposed sub-question is used to retrieve semantically similar AQGs through dense retrieval, which are then mapped to the source chunks. These chunks are reranked with respect to the original query using a cross-encoder to produce the final evidence set.
    \item \textbf{(4) Indexing and Inference Workflow.} The entire pipeline is structured into an offline indexing stage - where documents are transformed into AQG embeddings - and an online inference stage, where decomposed questions drive retrieval and generation.
\end{itemize}

\subsection{Query Decomposition with LLMs}
Multihop questions inherently involve implicit, multi-step reasoning over diverse entities, events, and facts. When complex queries are reduced to a single vector, the nuanced reasoning behind them often gets lost, impairing retrieval performance. To overcome this, we decompose each multihop question into a sequence of single-hop subquestions.

We leverage large instruction-tuned language models to perform this decomposition in a zero-shot setting. Specifically, we experimented with GPT-4o (OpenAI 2024), LLaMA-3.1-8B-inst (Abhimanyu Dubey et al. 2024), and Gemini-1.5-Pro (Gemini Team Google et al. 2024). Given a multi-hop input question $q$, each model is prompted to generate a list $\{q_1, q_2, \ldots, q_n\}$ of logically intermediate single-hop subquestions that correspond to steps in the reasoning chain.

This decomposition has the following properties:
\begin{itemize}
    \item \textbf{Semantically aligned:} Each subquestion closely reflects the intent of the original question, focusing on a single factual point.
    \item \textbf{Efficient and zero-shot:} It does not require specialized fine-tuning or labeled data.
    \item \textbf{Retrieval-friendly:} Subquestions are phrased to maximize retrievability from our document-side AQG index.
\end{itemize}

This LLM-driven decomposition makes it possible to isolate individual reasoning steps, increasing retrieval coverage, and reducing ambiguity in multi-hop query matching.

\subsection{Document-side Answerable Question Generation}
Conventional Retrieval-Augmented Generation systems commonly employ dense encoders to directly embed document chunks. However, this method encounters a structural discrepancy: queries are characteristically succinct and interrogative in form, whereas documents are predominantly expansive and expository. This disparity leads to representational divergence within the embedding space.

To mitigate this, we introduce a document-side transformation that converts each document chunk into a set of answerable questions (AQGs). These serve as surrogates for the original chunk, formulated to closely emulate the manner in which users typically pose inquiries.

We employ the Qwen3-8B model to generate answerable questions from each chunk $d_i$. Specifically, for each chunk, we prompt the model to generate $m$ questions $\{a_{i1}, a_{i2}, \ldots, a_{im}\}$ that can be directly answered from $d_i$ without requiring external context. On average, the model produces approximately 10 questions per chunk. For a detailed breakdown of answerable question generation statistics and their distribution across datasets, please refer to Appendix~\ref{sec:aqg-appendix}.

Each answerable question is then embedded using the \texttt{multilingual-e5-large} encoder~\cite{wang2024multilingual}, the same model used for embedding user queries. These AQG embeddings are indexed and mapped to their originating document chunk.

\subsubsection*{Document Chunking}
To prepare documents for AQG generation, we segment each source document into overlapping chunks using a sliding window of 800 characters with a stride of 600 characters, resulting in 200-character overlaps. This chunking strategy ensures sufficient context is preserved across chunk boundaries, which is critical for generating coherent and self-contained answerable questions. It also balances retrieval granularity and coverage, allowing downstream models to identify precise yet context-rich evidence spans.

\begin{algorithm}[t]
\caption{Answerable Question Guided Retrieval-Augmented Generation}
\label{alg:aqgrag}
\begin{algorithmic}[1]
\REQUIRE Multihop question $q$, AQG index $\mathcal{V}$ (maps answerable question $\rightarrow$ source document chunk), encoder $E$, reranker $R$, generator $G$, parameters $k_1$, $k_2$
\ENSURE Final generated answer

\STATE \textbf{Step 1: Question Decomposition}
\STATE Use a pretrained LLM to decompose $q$ into single-hop subquestions $\{q_1, q_2, ..., q_n\}$

\STATE \textbf{Step 2: AQG-Based Retrieval}
\STATE Initialize set $\mathcal{A} \gets \emptyset$
\FOR{each $q_i$ in $\{q_1, ..., q_n\}$}
    \STATE Embed $q_i$ using encoder $E$
    \STATE Retrieve top-$k_1$ most similar answerable questions from $\mathcal{V}$
    \STATE Add retrieved (AQG, source chunk) pairs to $\mathcal{A}$
\ENDFOR

\STATE \textbf{Step 3: Candidate Chunk Collection}
\STATE Sort $\mathcal{A}$ by similarity to $q_i$ and collect associated document chunks
\STATE Deduplicate chunks to form candidate set $\mathcal{D}$

\STATE \textbf{Step 4: Cross-Encoder Reranking}
\FOR{each chunk $d \in \mathcal{D}$}
    \STATE Compute relevance score $s_{q,d}$ using reranker $R(q, d)$
\ENDFOR
\STATE Select top-$k_2$ chunks with highest $s_{q,d}$

\STATE \textbf{Step 5: Answer Generation}
\STATE Concatenate top-$k_2$ chunks into context $C$
\STATE Generate answer $a \gets G(q, C)$

\RETURN $a$
\end{algorithmic}
\end{algorithm}

\subsection{Two-Stage Retrieval and Reranking}
Our pipeline adopts a two-stage retrieval strategy to effectively balance recall and precision when handling decomposed subquestions and complex multihop queries.

\paragraph{Stage 1: AQG-based Retrieval.}
For each decomposed single-hop query $q_i$, we embed it using the E5 encoder and perform dense retrieval over the AQG embedding index constructed from all answerable questions. Each AQG is linked to its source document chunk $d_j$, enabling us to retrieve relevant evidence indirectly via semantically aligned subquestions.

We sort AQGs by their similarity to the query, and collect the corresponding source chunks. Since multiple AQGs may point to the same document, this process yields a high-quality document candidate pool with natural deduplication.

\paragraph{Stage 2: Original-Question Reranking.}
We then rerank the deduplicated document set using a cross-encoder that scores the relevance of each candidate chunk $d_j$ with respect to the original multihop question $q$. This reranking filters out contextually irrelevant documents that may have been retrieved via valid but tangential subquestions. The final top-ranked documents are passed to the generator for answer synthesis.

\subsection{Indexing and Inference Workflow}
Our system is composed of two main workflows: an offline indexing phase where document representations are constructed, and an online inference phase where multihop queries are processed and answers are generated.

\paragraph{Offline Indexing.}
Given a corpus of documents, we first segment them into overlapping chunks using a fixed window size. Each chunk $d_i$ is passed through the Qwen3-8B model to generate a set of $m$ answerable questions $\{a_{i1}, a_{i2}, \ldots, a_{im}\}$, which can be answered based solely on the content of $d_i$. These questions are then embedded using the same retriever model employed during inference (e.g., E5). We index these AQG embeddings in a vector database, each entry pointing back to its originating chunk. This index serves as the document-side retrieval base and is computed entirely offline.

\paragraph{Online Inference.}
At inference time, a user submits a multihop question $q$. A decomposition model (e.g., GPT-4o, Gemini-1.5-pro, or LLaMA-3.1-8B-inst) segments $q$ into single-hop subquestions $\{q_1, \ldots, q_n\}$. Each $q_i$ is embedded and used to query the AQG index for top-$k_1$ matches, which are mapped back to their source document chunks. These candidate chunks are then reranked using a cross-encoder that compares each chunk $d$ to the original complex query $q$. The top-$k_2$ reranked chunks are selected and passed to a generator model, which produces the final answer.

This separation of indexing and inference enables efficient, reusable retrieval while ensuring semantically aligned generation without online paraphrasing or summarization.

\section{Experiments}

\begin{table*}[t]
\centering
\begin{tabular}{lcccc}
\toprule
\rowcolor{gray!20}
\textbf{Model} & \textbf{2WikiMultiHopQa} & \textbf{HotpotQA} & \textbf{MuSiQue} & \textbf{Average} \\
\midrule

\addlinespace
\rowcolor{gray!10}
\multicolumn{5}{c}{\textbf{\textit{Without RAG}}} \\
\addlinespace
Gemini-1.5-pro          & 38.31 & 36.79 & 20.09 & 31.73 \\
GPT-4o                  & 40.62 & 46.76 & 30.76 & 39.38 \\

\addlinespace
\rowcolor{gray!10}
\multicolumn{5}{c}{\textbf{\textit{RAG with ROT}}} \\
\addlinespace
Self-RAG                & 46.75 & 50.51 & 24.62 & 40.63 \\
Query Rewriting         & - & 43.85 & - & 43.85 \\
ReSP                    & 38.3 & 47.2 & - & 42.75 \\

\addlinespace
\rowcolor{gray!10}
\multicolumn{5}{c}{\textbf{\textit{RAG with Reranking}}} \\
\addlinespace

GPT-3.5-Turbo           & 43.44 & 52.31 & 25.22 & 40.32 \\

\addlinespace
\multicolumn{5}{c}{\textbf{\textit{LLaMA-3.1-8B-inst}}} \\
\addlinespace
\rowcolor{blue!5}
LLaMA-3.1-8B-inst       & 46.33 & 52.50 & 26.70 & 41.84 \\
\rowcolor{blue!5}
RAPTOR (LLaMA-3.1-8B-inst)   & 43.61 & 52.30 & 23.79 & 39.90 \\
\rowcolor{blue!5}
MacRAG (LLaMA-3.1-8B-inst)       & 44.87 & \textbf{57.39} & 30.38 & 44.21 \\
\rowcolor{blue!5}
\textbf{Ours (LLaMA-3.1-8B-inst)} & \textbf{50.49} & 56.42 & \textbf{33.81} & \textbf{46.91} \\

\addlinespace
\multicolumn{5}{c}{\textbf{\textit{GPT-4o}}} \\
\addlinespace
\rowcolor{cyan!8}
LongRAG (GPT-4o)        & 59.97 & 65.46 & 38.98 & 54.80 \\
\rowcolor{cyan!8}
MacRAG (GPT-4o)         & 59.00 & 67.15 & 44.76 & 56.97 \\
\rowcolor{cyan!8}
\textbf{Ours (GPT-4o)}  & \textbf{69.19} & \textbf{67.50} & \textbf{54.99} & \textbf{63.89} \\

\addlinespace
\multicolumn{5}{c}{\textbf{\textit{Gemini-1.5-pro}}} \\
\addlinespace
\rowcolor{purple!6}
LongRAG (Gemini-1.5-pro)        & 60.13 & 63.59 & 34.90 & 52.87 \\
\rowcolor{purple!6}
MacRAG (Gemini-1.5-pro)         & 58.38 & 63.02 & 43.31 & 54.90 \\
\rowcolor{purple!6}
\textbf{Ours (Gemini-1.5-pro)}  & \textbf{63.61} & \textbf{66.85} & \textbf{48.44} & \textbf{59.63} \\

\bottomrule
\end{tabular}
\vspace{0.5em}
\caption{F1-score comparisons with other methods on HotpotQA, 2WikimultihopQA, and MuSiQue are drawn from LongBench~\cite{bai2024longbenchbilingualmultitaskbenchmark}. Our method achieves superior F1 performance on multihop QA datasets, consistently outperforming baselines across most model types and datasets.}
\label{tab:main_results}
\end{table*}

\subsection{Experimental Setup}

We evaluate our approach on three challenging multi-hop question answering benchmarks from LongBench~\cite{bai2024longbenchbilingualmultitaskbenchmark}: HotpotQA~\cite{yang2018HotpotQA}, 2WikiMultiHopQa~\cite{ho2020constructing}, and MuSiQue~\cite{trivedi2022MuSiQue}. These datasets are designed to test a system’s ability to retrieve and reason across disjoint evidence sources, posing challenges such as the “Lost in the Middle” effect and semantic drift. To simulate open-domain settings, we use the full-wiki setup for all tasks.

\paragraph{Models and Evaluations.} Our evaluations employ three recent LLMs: GPT-4o~\cite{hurst2024gpt}, Gemini-1.5-pro~\cite{team2024gemini}, and the open-source LLaMA-3.1-8B-instruct~\cite{dubey2024llama}. Performance is measured using F1-score. We compare our model primarily against the following baselines:

\begin{itemize}
    \item \textbf{No-RAG LLMs:} Each LLM directly answers questions without retrieval.
    \item \textbf{RAG with reasoning-oriented transformations (ROT):} Self-RAG~\cite{gao2023selfrag} a generation-side control method, enables multi-step reasoning by dynamically deciding when to retrieve and how to evaluate retrieved content using reflection tokens. Query Rewriting~\cite{ma2023query}, a \textit{question-side transformation} method, reformulates queries to better match retrievable content. ReSP~\cite{jiang2025retrieve}, a \textit{document-side transformation} approach, iteratively summarizes retrieved passages and plans the next retrieval step. Each method supports multi-step reasoning through distinct transformation strategies across the retrieval and generation pipeline.
    \item \textbf{RAG with reranking:} Standard dense retrieval pipelines with cross-encoder reranking, including RAPTOR~\cite{sarthi2024raptor} and LongRAG~\cite{jiang2024longrag} and MacRAG~\cite{lim2025macrag}, which serves as a strong recent baseline. MacRAG adopts the same LLM and evaluates on the same three datasets as used in our work, enabling a direct and fair comparison. Importantly, MacRAG also conducts experiments with LongRAG under the identical setting, and we incorporate those results for direct comparison in this paper.
\end{itemize}

We evaluate performance using the standard F1 score, averaged across all datasets. For each subquestion, we retrieve the top $k_1 = 100$ AQG candidates and rerank them to select the top $k_2 = 7$ document chunks for final answer generation. We use \texttt{ms-macro-MiniLM-L-12-v2}, the same cross-encoder reranker used in MacRAG, throughout our system.

Unlike MacRAG and LongRAG, which compares multiple generation modes (e.g., R\&L, Full E\&F), we adopt a single-generation scheme: top-$k_2$ reranked chunks are passed directly to the LLM for single-step generation (R\&B). We adopt this setup to better isolate the impact of retrieval and reranking, without the added variability introduced by complex generation strategies.

\subsection{Ablation and Analysis Overview}

We conduct a series of ablation studies and detailed analyses to investigate the contribution of each component in our framework and to better understand design choices.

\paragraph{Document vs. AQG Embeddings.}  
To evaluate the effectiveness of using answerable questions (AQGs) as document-side proxies, we compare three retrieval configurations over the same set of document chunks: (1) document-only embeddings, (2) AQG-only embeddings, and (3) a combination of both. This comparison allows us to quantitatively assess the performance contribution of embedding AQGs alone.

\paragraph{Effect of Query Decomposition.}  
To assess the impact of multihop query decomposition, we compare two retrieval strategies: one that uses the original multihop query and another that relies on its decomposed subquestions. This comparison not only quantifies the retrieval gains from decomposition, but also demonstrates the importance of constructing semantically independent subquestions. It suggests that having meaningfully distinct subqueries is beneficial not just from the document side, but also from the perspective of question formulation itself.

\subsection{Further Analyses}

We conduct a series of analyses to further understand the behavior and effectiveness of our proposed retrieval strategy under different settings:

\begin{itemize}
    \item \textbf{Reranking Scope ($k_2$).} We vary the number of reranked documents ($k_2 \in \{5, 7, 10, 12, 15\}$) to study how retrieval depth affects answer quality across models and datasets.

    \item \textbf{Decomposed Question Answering Strategy.} We compare two inference strategies: (a) sequential answering of decomposed subquestions and (b) unified answering using the original multihop question over the full retrieved context, to evaluate how reasoning style affects performance.

    \item \textbf{Summary and Paraphrase Embedding RAG.} For each document chunk used in AQG, we generate summary and paraphrased variants using a LLM. These alternative representations are embedded and used for retrieval to examine whether textual compression or transformation enhances semantic alignment in the RAG setting.
\end{itemize}

\subsection{Main Results}

Table~\ref{tab:main_results} presents F1 scores across three multihop QA datasets—2WikiMultiHopQa, HotpotQA, and MuSiQue—using Llama-3.1-8B-inst, GPT-4o, and Gemini-1.5-Pro. Our method consistently outperforms closed-book LLMs, baseline RAGs, and reranking-based RAG baselines like RAPTOR, LongRAG, and MacRAG.

Notably, with GPT-4o, our approach achieves the highest average F1 (63.89), surpassing MacRAG by 6.92 points. Similar gains are observed with Llama-3.1-8B-inst and Gemini-1.5-pro models. These results underscore the impact of our question-centric design: decomposing queries and indexing documents via AQGs improves semantic alignment and retrieval quality. Crucially, generating answerable questions compels the LLM to interpret each document chunk from multiple perspectives, enriching it into a more informative and retrieval-ready representation—thereby enhancing downstream multihop reasoning.

\begin{table}[t]
\centering
\captionsetup{font=small}
\caption{Ablation study results comparing different embedding sources—raw document chunks, automatically generated questions (AQG), and their combination—across the 2WikimultiHopQA, HotpotQA, and MuSiQue datasets. F1-scores are reported for GPT-4o, Gemini-1.5-pro, and LLaMA-3.1-8B-instruct.}
\small 
\resizebox{\linewidth}{!}{%
\begin{tabular}{llccc}
\toprule
\rowcolor{gray!20}
\textbf{Mode} & \textbf{Dataset} & \textbf{Document} & \textbf{AQG} & \textbf{Both} \\
\midrule
\multirow{3}{*}{GPT-4o} 
  & 2wikimulti & 38.95 & 67.29 & \textbf{69.19} \\
  & HotpotQA   & 44.37 & 65.69 & \textbf{67.50} \\
  & MuSiQue    & 29.26 & 54.08 & \textbf{54.99} \\
\midrule
\multirow{3}{*}{Gemini-1.5-pro}
  & 2wikimulti & 25.48 & 66.82 & \textbf{66.85} \\
  & HotpotQA   & 28.84 & 62.37 & \textbf{63.61} \\
  & MuSiQue    & 14.82 & 46.24 & \textbf{48.44} \\
\midrule
\multirow{3}{*}{LLaMA-3.1-8B-inst}
  & 2wikimulti & 18.05 & 49.40 & \textbf{50.49} \\
  & HotpotQA   & 22.04 & 56.31 & \textbf{56.42} \\
  & MuSiQue    &  7.92 & 30.50 & \textbf{33.81} \\
\bottomrule
\end{tabular}
}
\label{tab:embedding_ablation}
\end{table}

\subsection{Ablation: Embedding Source}

Table~\ref{tab:embedding_ablation} shows an ablation study examining the impact of different indexing strategies: (1) directly embedding document chunks, (2) embedding answerable questions (AQGs), and (3) using both representations together.

Across all models and datasets, using AQGs alone leads to substantial performance gains compared to using raw document chunks. For instance, on MuSiQue, GPT-4o improves from 29.26 to 54.08 when switching from document embeddings to AQG embeddings. Similar patterns are observed for Gemini-1.5-Pro (14.82 $\rightarrow$ 46.24) and LLaMA-3.1-8B-inst (7.92 $\rightarrow$ 30.50).

Adding document embeddings on top of AQGs yields marginal improvements in some cases. For example, on 2WikiMultiHopQa with GPT-4o, performance increases from 67.29 to 69.19. However, in most cases, the difference between AQG-only and Both is small (often less than 2 points), suggesting that AQG representations capture most of the essential semantic content for retrieval.

The performance gap between document-only and AQG-based indexing was larger than expected.
This suggests that AQG-based representations are not only structurally better aligned with natural language queries, but also more effective at capturing and expressing the essential information needed for retrieval.

\subsection{Ablation: Query Decomposition}

We investigate the impact of question decomposition by comparing two inference setups: (1) using the original multihop question as a single query, and (2) using multiple decomposed subquestions with answerable question–guided retrieval. Table~\ref{tab:decomposition_ablation} summarizes the results.

\begin{table}[h]
\centering
\caption{F1-score comparison between decomposed and original multihop queries across the 2WikiMultiHopQA, HotpotQA, and MuSiQue datasets. Results are shown for GPT-4o, Gemini-1.5-pro, and LLaMA-3.1-8B-instruct models.}
\label{tab:decomposition_ablation}
\small
\resizebox{\columnwidth}{!}{%
\begin{tabular}{llcc}
\toprule
\rowcolor{gray!20}
\textbf{Model} & \textbf{Dataset} & \textbf{Decomposed} & \textbf{Original} \\
\midrule
\multirow{3}{*}{GPT-4o} 
  & 2WikiMultiHopQa & \textbf{69.19} & 63.17 \\
  & HotpotQA         & 67.50         & \textbf{68.35} \\
  & MuSiQue          & \textbf{54.99} & 47.87 \\
\midrule
\multirow{3}{*}{Gemini-1.5-pro} 
  & 2WikiMultiHopQa & \textbf{66.85} & 64.07 \\
  & HotpotQA         & 63.61         & \textbf{65.40} \\
  & MuSiQue          & \textbf{48.44} & 42.56 \\
\midrule
\multirow{3}{*}{LLaMA-3.1-8B-inst} 
  & 2WikiMultiHopQa & \textbf{50.49} & 48.64 \\
  & HotpotQA         & \textbf{56.42} & 55.59 \\
  & MuSiQue          & \textbf{33.81} & 28.10 \\
\bottomrule
\end{tabular}
}
\end{table}

Across most cases, the decomposed queries outperform the original ones—particularly on datasets like 2WikiMultiHopQa and MuSiQue. In contrast, for HotpotQA, original queries yield slightly better performance when using GPT-4o and Gemini. This is likely because HotpotQA’s relatively simpler structure makes it more suitable for end-to-end holistic reasoning, reducing the benefit of decomposition. This pattern is also reflected in the decomposition statistics presented in Appendix~\ref{sec:appendix-decomq}.

\begin{figure*}[t]
    \centering
    \includegraphics[width=\textwidth]{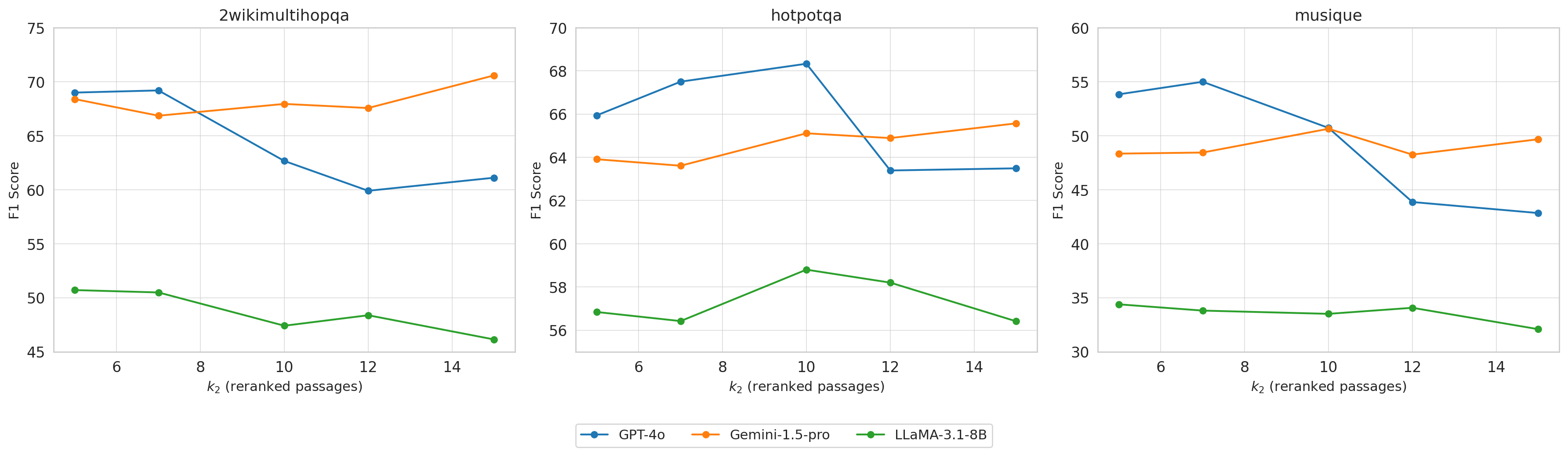}
    \caption{Effect of reranking scope ($k_2$) on multihop QA performance across datasets.}
    \label{fig:k2_ablation}
\end{figure*}

\paragraph{Reranking Scope ($k_2$).}
To analyze the effect of reranking scope, we vary the number of top-$k_2$ documents used in the final reranked retrieval step. As shown in Figure~\ref{fig:k2_ablation}, performance trends vary across datasets and model backbones.

GPT-4o performs best with moderate $k_2$ values, peaking at $k_2 = 7$ on 2WikiMultiHopQa and $k_2 = 10$ on HotpotQA, but degrades noticeably with larger $k_2$, especially on MuSiQue.

Gemini-1.5-Pro shows more stable or even improving performance as $k_2$ increases, particularly on 2WikiMultiHopQa and HotpotQA, peaking around $k_2 = 15$.

LLaMA-3.1-8B-inst on the other hand, consistently suffers from performance drops as $k_2$ increases, indicating difficulties in leveraging longer context windows effectively.

These results suggest that the optimal number of reranked passages ($k_2$) is model- and dataset-dependent. Smaller $k_2$ values are generally safer when using models that struggle with long context (e.g., LLaMA-3.1-8B-inst), while larger values benefit stronger models like Gemini in scenarios with informative passages.

\paragraph{Sequential vs. Unified Inference.}
Our experiments compare sequential answering over decomposed subquestions with unified inference using the original query and full retrieved context.(Table ~\ref{tab:seq_vs_unified}) The results demonstrate that unified inference consistently outperforms sequential generation for large-scale models such as GPT-4o and Gemini. The performance gains are particularly pronounced on datasets with high inter-document dependency, such as 2WikiMultiHopQa and MuSiQue, indicating that reasoning over a fully integrated context enables more effective multi-hop answer synthesis.

These findings underscore the importance of post-retrieval reasoning strategy and suggest that unified inference provides a more robust mechanism for information integration than aggregating intermediate answers. Interestingly, our results align with observations made in prior work such as LongRAG, which reports strong performance by leveraging long-context models to reason over retrieved document blocks as a whole. While our setup differs in architecture and retrieval granularity, the underlying conclusion remains consistent: models benefit from reasoning holistically over relevant information when capacity permits.

\begin{table}[h]
\centering
\caption{Comparison of sequential vs. unified inference methods over decomposed questions.}
\label{tab:seq_vs_unified}
\small
\resizebox{\columnwidth}{!}{%
\begin{tabular}{llcc}
\toprule
\rowcolor{gray!20}
\textbf{Model} & \textbf{Dataset} & \textbf{Sequential} & \textbf{Unified} \\
\midrule
\multirow{3}{*}{GPT-4o}
  & 2WikiMultiHopQa & 64.42 & \textbf{69.19} \\
  & HotpotQA        & 66.23 & \textbf{67.50} \\
  & MuSiQue         & 47.55 & \textbf{54.99} \\
\midrule
\multirow{3}{*}{Gemini-1.5-pro}
  & 2WikiMultiHopQa & 65.56 & \textbf{66.85} \\
  & HotpotQA        & 63.26 & \textbf{63.61} \\
  & MuSiQue         & 48.32 & \textbf{48.44} \\
\midrule
\multirow{3}{*}{LLaMA-3.1-8B-inst}
  & 2WikiMultiHopQa & \textbf{52.07} & 50.49 \\
  & HotpotQA        & 54.79 & \textbf{56.42} \\
  & MuSiQue         & 33.16 & \textbf{33.81} \\
\bottomrule
\end{tabular}
}
\end{table}

\paragraph{Summary and Paraphrase Embedding RAG.}
To assess the utility of transformed document content for retrieval, we compare four types of representations: the original document, summaries, paraphrased versions, and AQG-derived questions. For summarization and paraphrasing, we prompt an LLM to generate summaries or paraphrased versions of each document chunk individually.

\begin{figure}[t]
  \centering
  \includegraphics[width=\linewidth]{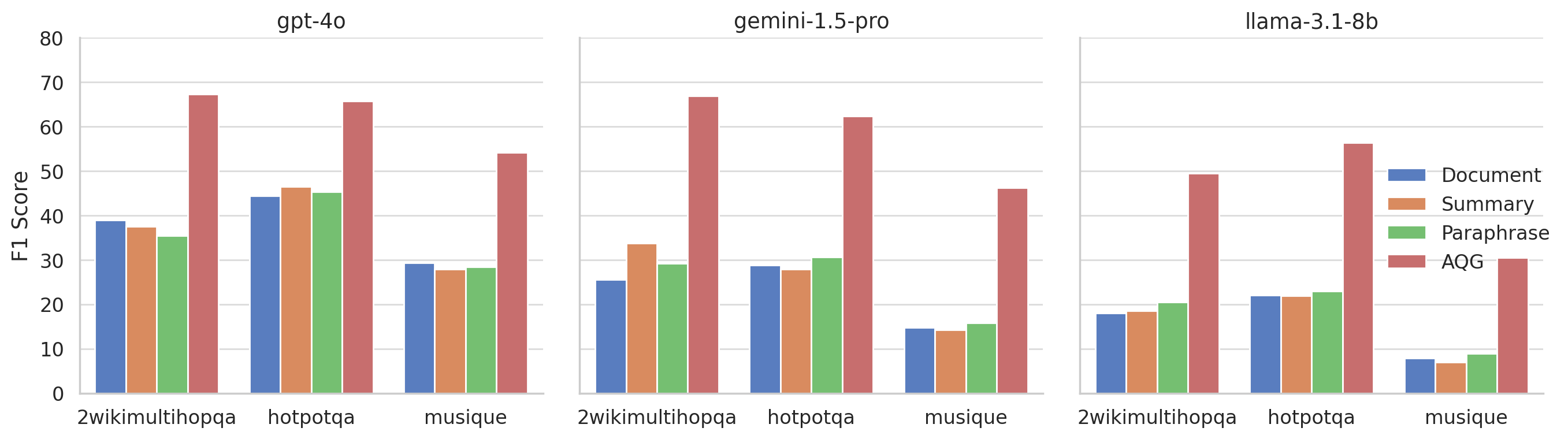}
  \caption{Comparison of F1 scores using different embedding strategies across datasets and LLMs: direct document embeddings, summary-based embeddings, paraphrase-based embeddings, and AQG-based embeddings. AQG outperforms other methods consistently.}
  \label{fig:summary_paraphrase_rag}
\end{figure}

As shown in Figure~\ref{fig:summary_paraphrase_rag}, both summarization and paraphrasing yield only marginal improvements over the raw document representation, and in some cases even result in performance degradation, depending on the model and dataset. Paraphrasing tends to slightly outperform summarization, which may be attributed to the fact that summarization often omits or compresses original information, while paraphrasing preserves the full content by merely rephrasing it. We provide all prompt templates used for AQ generation, summarization, and paraphrasing in Appendix~\ref{sec:prompt} to support reproducibility.

In contrast, AQG-derived queries lead to consistent and substantial improvements across all models and datasets. This suggests that query-oriented transformations offer stronger retrieval signals than document-side reformulations. The benefit becomes especially clear when using large-scale models capable of aligning the query and context semantically during generation.

\section{Conclusion}

In this work, we presented a novel, question-centric RAG framework that enhances multihop question answering by structurally aligning both the query and document sides of the retrieval pipeline. Our method leverages large language models to decompose complex multihop questions into semantically focused subquestions, and transforms document chunks into answerable question (AQ) representations. This dual transformation reduces the semantic-structural mismatch common in embedding-based retrieval systems and enables more effective information retrieval.

We evaluated our method across three established multihop QA benchmarks—HotpotQA, 2WikiMultiHopQa, and MuSiQue—and consistently observed strong performance gains over competitive baselines under various large model settings. Ablation analyses further show that AQ-based document indexing significantly contributes to retrieval effectiveness, and that question decomposition meaningfully restructures complex queries to better support downstream generation.

These findings suggest that representing documents through the lens of answerable questions is not only more compatible with the nature of question answering, but also improves the semantic accessibility and utility of large textual corpora.

\section{Limitations}

While our framework achieves strong performance, it comes with certain limitations. First, the use of multiple subquestions increases the inference latency due to repeated retrieval and reranking steps. Second, the quality of generated AQGs and their paraphrases or summaries depends on the underlying language model and may introduce noise. Lastly, although our experiments cover multiple datasets, further validation on other domains and languages is necessary to generalize our findings.

\bibliography{aaai2026}

\appendix

\section{Analysis of Decomposed Question Distribution.} \label{sec:appendix-decomq}
\begin{figure}[h]
    \centering
    \includegraphics[width=\columnwidth]{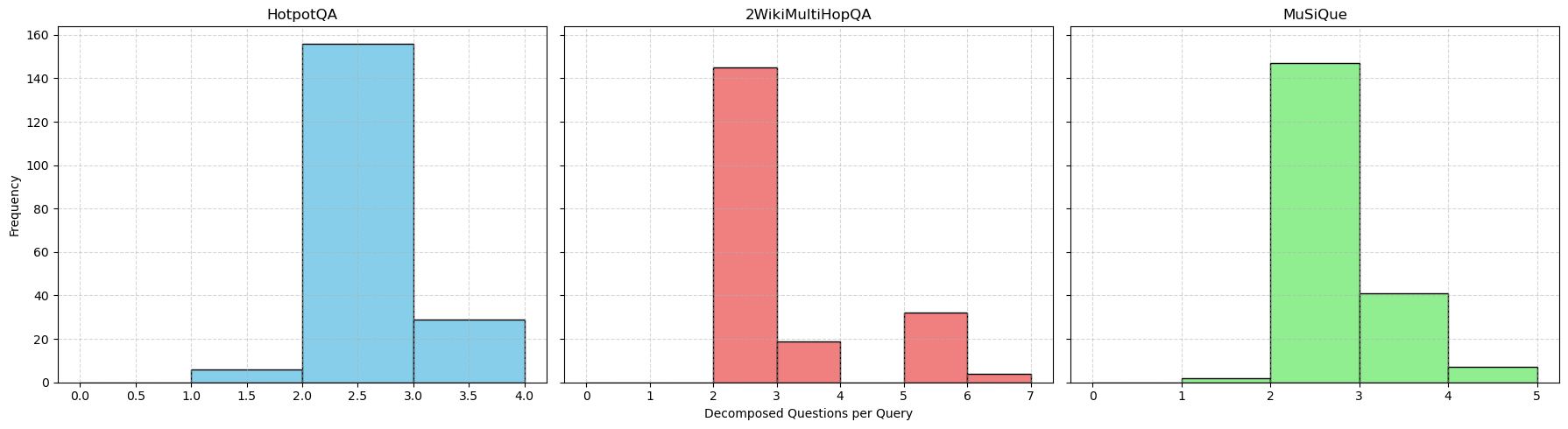}
    \caption{
    Histogram showing the number of decomposed questions per original query across test datasets.
    }
    \label{fig:decomp-histogram}
\end{figure}

As shown in Table~\ref{tab:decomp_stats} and Figure~\ref{fig:decomp-histogram}, 
among the three datasets, HotpotQA exhibits the narrowest distribution of decomposed question counts, 
indicating that its queries were decomposed into relatively fewer sub-questions compared to the others. This observation supports the finding from our query decomposition ablation study, where the performance difference between models with and without decomposition was smallest for HotpotQA.

\begin{table}[htbp]
\centering
\small
\begin{tabular}{lcccc}
\toprule
\textbf{Dataset} & \textbf{Mean} & \textbf{Std Dev} & \textbf{Min} & \textbf{Max} \\
\midrule
HotpotQA         & 2.21  & 0.58  & 1 & 5 \\
2WikiMultiHopQA   & 2.66  & 1.20  & 2 & 6 \\
MuSiQue          & 2.31  & 0.63  & 1 & 5 \\
\bottomrule
\end{tabular}
\caption{
Statistics of the number of decomposed questions per original query across test datasets.
}
\label{tab:decomp_stats}
\end{table}

\section{Analysis of Question Generation Distribution.} \label{sec:aqg-appendix}
\begin{figure}[h]
    \centering
    \includegraphics[width=\columnwidth]{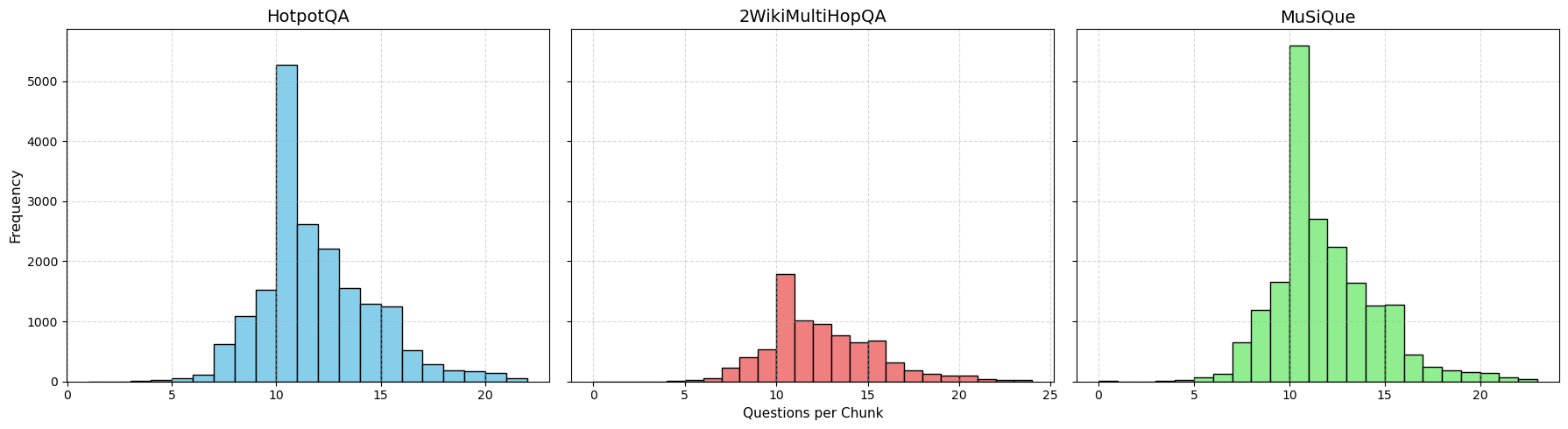}
    \caption{
    Histogram of generated question counts per chunk in HotpotQA, 2WikiMultiHopQA, and MuSiQue.}
    \label{fig:aqg-histogram}
\end{figure}

As shown in Table~\ref{tab:aq_stats} and Figure~\ref{fig:aqg-histogram}, 
the number of answerable questions generated per chunk remains consistent across all three datasets, 
with average values around 11--12 and standard deviations close to 3.5. 
The histogram illustrates that most chunks fall within a narrow band of approximately 8 to 15 questions. 
In the case of MuSiQue, a small number of chunks exhibit unusually high counts, 
resulting in a long-tailed distribution. 
Chunks with zero generated questions are typically composed solely of numeric or formulaic content, 
which does not lead to meaningful question generation.

\begin{table}[H]
\centering
\small
\begin{tabular}{lcccc}
\toprule
\textbf{Dataset} & \textbf{Mean} & \textbf{Std Dev} & \textbf{Min} & \textbf{Max} \\
\midrule
HotpotQA         & 11.59 & 3.34 & 0 & 83 \\
2WikiMultiHopQA   & 12.15 & 3.73 & 0 & 74 \\
MuSiQue          & 11.47 & 3.52 & 0 & 219 \\
\bottomrule
\end{tabular}
\caption{
Basic statistics of AQ-generated questions per chunk across datasets.
}
\label{tab:aq_stats}
\end{table}

\label{sec:appendix-aqg}

\section{Prompt Templates} \label{sec:prompt}
We include the prompt templates used to generate three types of document transformations: \textbf{Answerable Questions (AQs)}, \textbf{Summarizations}, and \textbf{Paraphrases}. All prompts were designed for use with Qwen3-8B. Each prompt takes a document chunk as input and returns a structured output tailored for different objectives:

\begin{itemize}
    \item \textbf{AQG Prompt:} This prompt instructs the LLM to generate multiple factual questions that can be answered directly from the input passage. The goal is to reformulate document content into a question-centric representation that supports better alignment with user queries during retrieval.

    \item \textbf{Summarization Prompt:} This prompt asks the model to generate a concise summary capturing all salient factual content in the passage. It is designed to preserve information while reducing length, enabling more efficient context input during inference.

    \item \textbf{Paraphrasing Prompt:} This prompt guides the model to rewrite the input passage using different phrasing and structure without omitting any key information. It aims to create semantically equivalent variants of the original content, preserving fidelity while increasing lexical diversity.
\end{itemize}

\subsection{Answerable Question Generation (AQG) Prompt} \label{sec:aqg_prompt}

We use the following instruction prompt to generate diverse, directly answerable questions from a document chunk. It guides the LLM to cover factual and conceptual information with explicit noun references and non-redundant phrasing.

\lstset{
  basicstyle=\ttfamily\footnotesize,
  breaklines=true,
  frame=single,
  backgroundcolor=\color{gray!10},
  columns=flexible
}
\vspace{1.0em}
\noindent
\begin{lstlisting}
# Role and Objective
You are a question generation assistant that analyzes a chunk of English text and generates a complete set of natural, clearly phrased questions that can be directly answered using only the information in that chunk.

# Instructions
- Accept a chunk of English text as input (not pre-split sentences).
- Analyze the entire chunk as a unified context.
- Generate a flat list of natural, well-formed English questions.
- Questions must:
  - Be directly answerable using only the given text.
  - Include a variety of types:
    - Factual (who, what, where, when)
    - Conceptual (why, how)
    - Summarizing or paraphrasing ("What is the main idea of the text?")
  - Use explicit noun references only.
  - Be non-redundant.

# Output Format
["Question 1", "Question 2",...]
\end{lstlisting}

\subsection{Summarization Prompt}
\label{sec:summarization_prompt}

We use the following instruction prompt to generate multiple concise and fact-preserving summaries from a document chunk. The LLM is guided to maintain content fidelity and create distinct summaries with varying phrasing.
\vspace{1.0em}
\lstset{
  basicstyle=\ttfamily\footnotesize,
  breaklines=true,
  frame=single,
  backgroundcolor=\color{gray!10},
  columns=flexible
}
\noindent
\begin{lstlisting}
# Role and Objective
You are a Summarization Assistant. Your task is to generate multiple accurate and concise summaries of a given English text. Each summary must capture the core information and key points of the original, written clearly and objectively.

# Instructions
- Generate up to ten distinct summaries for each input text.
- All summaries must convey the essential facts or ideas from the original.
- Focus on information-rich, content-preserving summaries - no opinions or interpretations.
- Do not invent, omit, or distort any important details from the original.
- Maintain a neutral, factual tone unless the source text dictates otherwise.
- Output each set of summaries as a Python list of strings.
- Ensure each summary version is meaningfully distinct in wording or structure, not just minor rephrases.

# Reasoning Steps / Workflow
1. Read and comprehend the full text (up to around 800 characters).
2. Identify the main message, key facts, and important supporting details.
3. Draft multiple summaries that concisely express these core points.
4. Eliminate versions that introduce factual errors or omit crucial content.
5. Present the final summaries as a Python list.

# Output Format
["Summary version 1", "Summary version 2", "Summary version 3", ...]

# JSON STRUCTURE INTEGRITY:
- Always return only pure JSON - no markdown, no extra text, no comments.
- All array elements must be separated by commas.
- All key-value pairs in objects must be separated by commas.
\end{lstlisting}

\subsection{Paraphrasing Prompt} \label{sec:paraphrasing_prompt}

This prompt guides the LLM to generate multiple paraphrased versions of a given text chunk while strictly preserving its original meaning, tone, and formality. Each paraphrase is distinct and semantically equivalent to the input.
\vspace{1.0em}
\lstset{
  basicstyle=\ttfamily\footnotesize,
  breaklines=true,
  frame=single,
  backgroundcolor=\color{gray!10},
  columns=flexible
}
\noindent
\begin{lstlisting}
# Role and Objective
You are a Paraphrasing Assistant. Your task is to generate multiple paraphrased versions of given English text chunks. Each paraphrase must express exactly the same meaning as the original, using different wording while keeping the tone and formality consistent.

# Instructions
- Generate up to ten distinct paraphrases for each input text chunk.
- Maintain the same tone and style as the original (e.g., formal/informal).
- DO NOT alter the meaning in any way. Preserving the exact original meaning is your top priority.
- If a paraphrase introduces even a subtle change in meaning, discard it.
- Output each paraphrased set as a Python list of strings.
- Ensure each string in the list is a distinct paraphrased version of the original text.
- Avoid overly repetitive or minimal variations. Aim for diverse and natural alternatives within the same meaning.

# Reasoning Steps / Workflow
1. Read the input text chunk.
2. Understand the precise meaning and tone of the sentence.
3. Brainstorm multiple alternative phrasings that retain the meaning and tone.
4. Filter out any versions that shift the meaning or nuance.
5. Present the final paraphrased versions as a Python list.

# Output Format
["Paraphrased version 1", "Paraphrased version 2", "Paraphrased version 3", ...]

# JSON STRUCTURE INTEGRITY:
- Always return only pure JSON - no markdown, no extra text, no comments.
- All array elements must be separated by commas.
- All key-value pairs in objects must be separated by commas.
\end{lstlisting}

\end{document}